\documentclass[letterpaper, 10 pt, conference]{ieeeconf}  

\IEEEoverridecommandlockouts                              

\usepackage{amsfonts}
\usepackage{amsmath}
\usepackage{amssymb}
\usepackage{authblk}
\usepackage{booktabs}
\usepackage{colortbl}
\usepackage{comment}
\usepackage{graphicx}
\usepackage{multicol}
\usepackage{subcaption}
\usepackage{tabularx}
\usepackage{hyperref}
\usepackage[normalem]{ulem}

\usepackage[ruled,vlined]{algorithm2e}
\usepackage[export]{adjustbox}
\usepackage{lscape}

\definecolor{lightgray}{rgb}{0.5, 0.5, 0.5}
\newcommand\graytext[1]{\textcolor{lightgray}{#1}}

\newcommand{\myurl}{\url{https://github.com/LostXine/crossway_diffusion}}

\title{\LARGE \bf
Crossway Diffusion: Improving Diffusion-based Visuomotor Policy\\ via Self-supervised Learning
}

\author{Xiang Li, Varun Belagali, Jinghuan Shang and Michael S. Ryoo
\thanks{The authors are with the Department of Computer Science, Stony Brook University, Stony Brook, New York, 11790. {\tt\small\{xiangli8, vbelagali, jishang, mryoo\}@cs.stonybrook.edu}}
\thanks{
The appendix, code, pretrained checkpoints, and datasets are available at \myurl.
}
}

\begin{document}
\maketitle
\thispagestyle{empty}
\pagestyle{empty}

\begin{abstract}

Sequence modeling approaches have shown promising results in robot imitation learning. 
Recently, diffusion models have been adopted for behavioral cloning in a sequence modeling fashion, benefiting from their exceptional capabilities in modeling complex data distributions. 
The standard diffusion-based policy iteratively generates action sequences from random noise conditioned on the input states. 
Nonetheless, the model for diffusion policy can be further improved in terms of visual representations.
In this work, we propose Crossway Diffusion, a simple yet effective method to enhance diffusion-based visuomotor policy learning via a carefully designed state decoder and an auxiliary self-supervised learning (SSL) objective.
The state decoder reconstructs raw image pixels and other state information from the intermediate representations of the reverse diffusion process.
The whole model is jointly optimized by the SSL objective and the original diffusion loss.
Our experiments demonstrate the effectiveness of Crossway Diffusion in various simulated and real-world robot tasks, confirming its consistent advantages over the standard diffusion-based policy and substantial improvements over the baselines.

\end{abstract}

\section{Introduction}

Behavioral Cloning (BC)~\cite{Pomerleau1988ALVINNAA} is a supervised learning formulation for robot action policy learning.
Given expert demonstration data consisting of a sequence of state-action pairs, a model is trained to predict the correct action vector given input states (e.g., images).
This framework has shown to be very effective particularly when a sufficient amount of training data is provided \cite{rt1}.

Recently, sequence modeling approaches~\cite{dt,trajectorytransformer,starformer} have been often used for behavioral cloning, because of their ability to model multiple steps of information. 
In such a formulation, the objective is to model the probability distribution of the multi-step state-action trajectory. 
This allows BC to consider beyond a single-step regression, better-taking advantage of history.
Given the success in modeling human language~\cite{devlin2018bert} and images~\cite{dosovitskiy2020image}, Transformers~\cite{att} have been popularly adopted for sequence modeling-based policies~\cite{dt,trajectorytransformer, starformer, rt1, rt2}. 

Diffusion models~\cite{sohl2015deep, ho2020denoising, ddim, burgert2023diffusion} have exceptional capabilities in modeling multimodal data distribution and generating new samples from the distribution, which make them suitable for imitating behaviors by generating trajectories. 
Recent works~\cite{janner2022planning, wang2022diffusion, pearce2023imitating} have successfully applied diffusion models for sequential modeling using low dimensional states.
For visuomotor control tasks, \cite{chi2023diffusion} demonstrated promising performance using multimodal states including visual observations as the conditions of the diffusion model.

In this work, we propose Crossway Diffusion, a simple yet effective method to enhance diffusion-based visuomotor policy learning via a carefully designed state decoder and a self-supervised learning (SSL) objective.
The state decoder reconstructs raw image pixels and other state information from the intermediate representations of the reverse diffusion process. 
The SSL objective and the original diffusion objective jointly optimize the whole model.

By doing so, we explicitly regularize the intermediate representation to capture the information of the input states, as now the model needs to reconstruct the input states from it.
Through our experiments over multiple challenging tasks from different benchmarks, we verify the consistent advantage of Crossway Diffusion in comparison to the baseline Diffusion Policy~\cite{chi2023diffusion}. 
Especially, our method achieves a $15.7\%$ improvement in success rate over the baseline on `Transport,~mh' dataset from Robomimic~\cite{mandlekar2021matters} and one variant of our method even achieves $17.1\%$, emphasizing the effectiveness of the SSL objective and the specific state decoder design.
Our contribution can be summarized as follows:
\begin{itemize}
    \item We propose Crossway Diffusion, which consistently improves diffusion-based visuomotor policy via a carefully designed state decoder and a simple SSL objective.
    \item We confirm the effectiveness of the proposed method on multiple challenging visual BC tasks from different benchmarks, including two real-world robot manipulation datasets we collected.
    \item We conduct detailed ablations on multiple design choices, verifying the advance and robustness of the proposed design.
\end{itemize}
\section{Preliminaries}
\subsection{Behavioral Cloning}
We consider simple behavioral cloning (BC) setting over a Markov Decision Process (MDP), described by the tuple $(\mathcal{S}, \mathcal{A}, P)$, where $s\in \mathcal{S}$ represents the state, $a\in \mathcal{A}$ is the action, and $P$ are the transition dynamics given by $P(s'|s, a)$.
A trajectory consists of a sequence of state-action pairs $\{s_0, a_0, s_1, a_1, \dots, s_T, a_T\}$, where $T$ is the length of sequence (task horizon).
Our goal is to train a robot policy $\pi$ that best recovers an unknown policy $\pi^*$ using a demonstration dataset $\mathcal{D} = \{ (s_i, a_i) \}$ collected by $\pi^*$. 
Specifically, the robot policy $\pi$ operates on a trajectory basis: $\pi(A_t | S_t )$, where $S_t=\{s_{t- T_s + 1}, s_{t - T_s + 2}, ..., s_t\}$ is the given short history state sequence, $A_t=\{a_{t}, a_{t + 1}, ..., a_{t + T_a-1}\}$ is the predicted future actions to take. $T_s$ and $T_a$ represent the lengths of these two sequences respectively. 

\subsection{Diffusion Models}
Diffusion models~\cite{sohl2015deep, ho2020denoising, ddim} are generative models that iteratively generate samples that match the data distribution.
Recent works~\cite{ramesh2022hierarchical, rombach2022high, gu2022vector, burgert2022peekaboo} have demonstrated its great ability in image generation and other data generation tasks. 
The diffusion process is of the original data being destroyed by a sequence of noise $q(x^k|x^{k-1})$ known as the forward process, where $k$ is the current diffusion step and there are $K$ steps in total. The diffusion model uses the reverse process, the backward process, $p_\theta(x^{k-1}|x^k)$ to denoise the corrupted data. 
By iteratively denoising, a diffusion model generates synthesized data $\hat{x}$ that approximates the original data distribution $q(x_0)$, starting from a random prior $p(x^K)$:
\begin{align}
    p_\theta(x^0) &= \int p_\theta(x^{0:K})dx^{1:K} \\
    &= \int p(x^K) \prod_{k=1}^{K}p_\theta(x^{k-1}|x^{k})dx^{1:K}
\end{align}
Typically the random prior $p(x^K)$ is a standard Gaussian distribution, and the denoising process is parameterized by the following Gaussian:
\begin{align}
    p_\theta(x^{k-1}|x^{k}) = \mathcal{N}(x^{k-1}|\mu_\theta(x^k,k),\Sigma^k),
\end{align}
where the parameter $\mu_\theta(x^k,k)$ is estimated from a neural network parameterized by $\theta$.

\subsection{Diffusion Models for Policy Learning}
Diffusion models have been applied for data augmentation~\cite{wang2023diffusion} and sample synthesis~\cite{lu2023synthetic, yu2023scaling, dai2023learning} in robot policy learning.
As recent works~\cite{dt,trajectorytransformer,shang2022starformer} formulate the robot policy learning as an action sequence generation problem, diffusion models have also been successfully adapted as sequence generation models~\cite{pearce2023imitating, wang2023diffusion, wang2022diffusion, janner2022planning}. 
Diffuser~\cite{janner2022planning} concatenates the low-dimensional states and the actions and models multiple state-action pairs as a matrix (image) generation problem. 
However, such a setting is not feasible for visuomotor policy learning due to the high dimensionality of the visual observations.

\begin{figure}[t]
    \centering
    \includegraphics[width=0.4\textwidth]{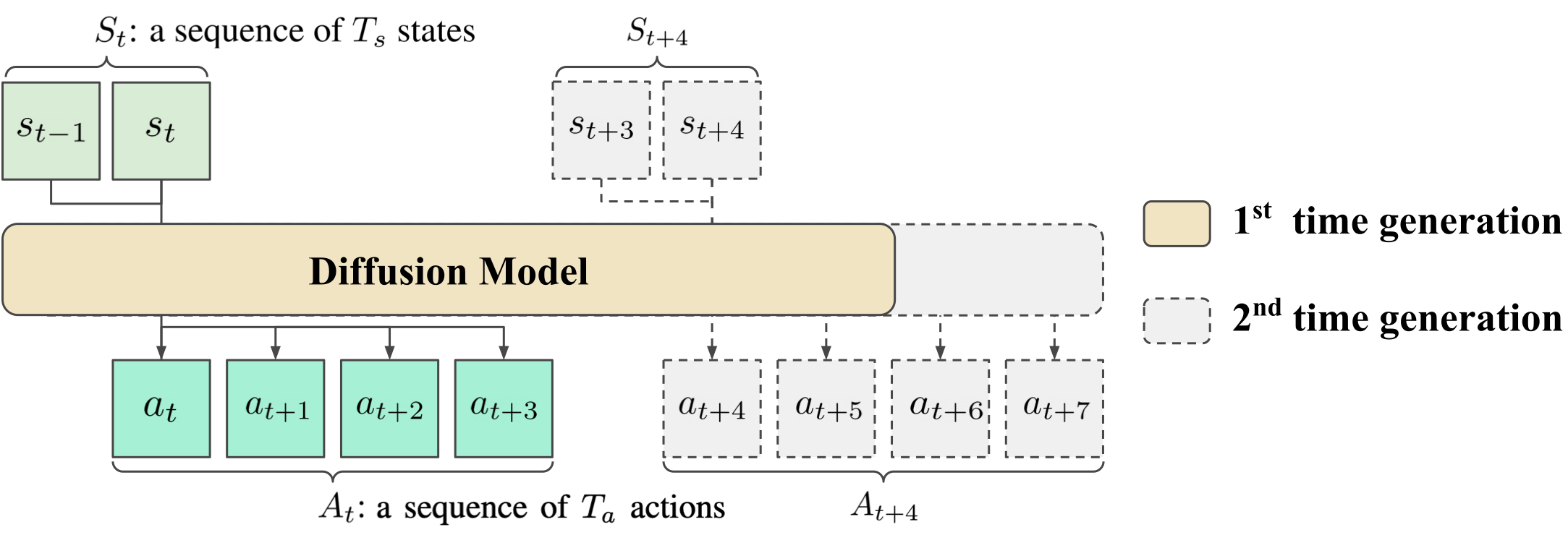}
    \caption{Trajectory generation formulation of Diffusion Policy. This figure shows a case where $T_s=2$ and $T_a=4$.}
    \label{fig:diffusion_policy}
\end{figure}

Diffusion Policy~\cite{chi2023diffusion} tackles the challenge by generating only action sequences, while conditioned on visual and other states.
Specifically, given a sequence of $T_s$ states $S_t=\{s_{t- T_s + 1}, s_{t - T_s + 2}, ..., s_t\}$ where each $s$ contains both visual states and low-dimensional states, the diffusion model generates a sequence of $T_a$ actions $A_t=\{a_{t}, a_{t + 1}, ..., a_{t + T_a-1}\}$ conditioned on the state sequence. 
When the agent finishes executing $A_t$, Diffusion Policy generates the consequent action sequence $A_{t + T_a}$ given $S_{t + T_a}$, which formulates a closed-loop control  (see Fig.~\ref{fig:diffusion_policy}).

The model of Diffusion Policy is composed of a state encoder~$E_S$, an action encoder~$E_A$, and an action decoder~$D_A$. 
The action encoder and decoder make up the diffusion model for generating action sequences by running the denoising process iteratively. 
The state encoder provides conditioning from the states, which modulates the generation process.
This paper focuses on the convolutional version of the Diffusion Policy due to its superior performance.

Given a state sequence with both visual and low-dimensional states $S_t=\{ S_{t,\textit{img}}, S_{t, \textit{low-dim}}\}$, the state encoder
extracts visual embeddings from images  $h_{t,\textit{img}}=E_S(S_{t,\textit{img}})$. 
The visual embeddings $h_{t,\textit{img}}$ are then concatenated with other low dimensional states $S_{t, \textit{low-dim}}$ to form the observation condition $h_t = h_{t, \textit{img}} \oplus S_{t, \textit{low-dim}}$. 

The action encoder $E_A$ takes the noisy action sequence $A_t^k$ at diffusion step $k$ and the observation condition $h_t$ and produces the representation $X_t^k$ from the deepest layer and other tensors for skip connection $X_{t, \textit{skip}}^k$ from the shallower layers: $X_t^k, X_{t, \textit{skip}}^k=E_A(A_t^k, h_t, k)$, 
where $t$ is the timestamp of a state or action trajectory. $X_t^k \in \mathbb{R}^{T \times C}$ where $T$ is the representation length along the time axis and $C$ is the number of channels.

The action decoder takes both $X_t^k$, $X_{t, \textit{skip}}^k$ and condition $h_t$ to estimate the noise $\epsilon$ which is applied to $A_t^k$ in the forward diffusion process.
The condition $h_t$ is applied between two convolutional layers in the residual block, using Feature-wise Linear Modulation (FiLM)~\cite{perez2018film}. 
Then a (slightly) denoised action sequence $A_t^{k-1}$ is derived from the estimated noise $\epsilon_\theta$ and $A_t^{k}$ using Eq.\ref{eq:unet}.
\begin{equation}
\begin{split}
    \epsilon_\theta&=D_A(X_t^k, X_{t, \textit{skip}}^k, h_t)\\
    A_t^{k-1}&=\frac{1}{\sqrt{\alpha_k}}\left( A_t^{k} - \frac{1-\alpha_k}{\sqrt{1-\bar{\alpha_k}}}\epsilon_\theta \right) + \sigma_kz
\end{split}
\label{eq:unet}
\end{equation}
where $z$ is randomly sampled from a normal distribution with the same dimension as $A_t^{k}$. 
$\alpha_k$, $\hat{\alpha_k}$, and $\sigma_k$ are parameters regarding the diffusion process used in DDPM~\cite{ho2020denoising}, except that we use $k$ as the diffusion step instead of $t$.

During inference, the denoising process mentioned above is repeated for $K$ times iteratively, generating a noiseless action sequence $A_t^0$ in the end.
The whole model is optimized using the same Mean Squared Error (MSE) as DDPM~\cite{ho2020denoising} to predict the noise $\epsilon$ applied to $A_t^0$ for constructing $A_t^k$.
\begin{equation}
    \mathcal{L}_{\textit{DDPM}} = \mathrm{MSE}(D_S(E_A(A_t^k, h_t, k), h_t), \epsilon)    
    \label{eq:ddpm}
\end{equation}

\section{Method}
\label{sec:method}
\begin{figure}[tb]
    \centering
    \includegraphics[width=0.485\textwidth]{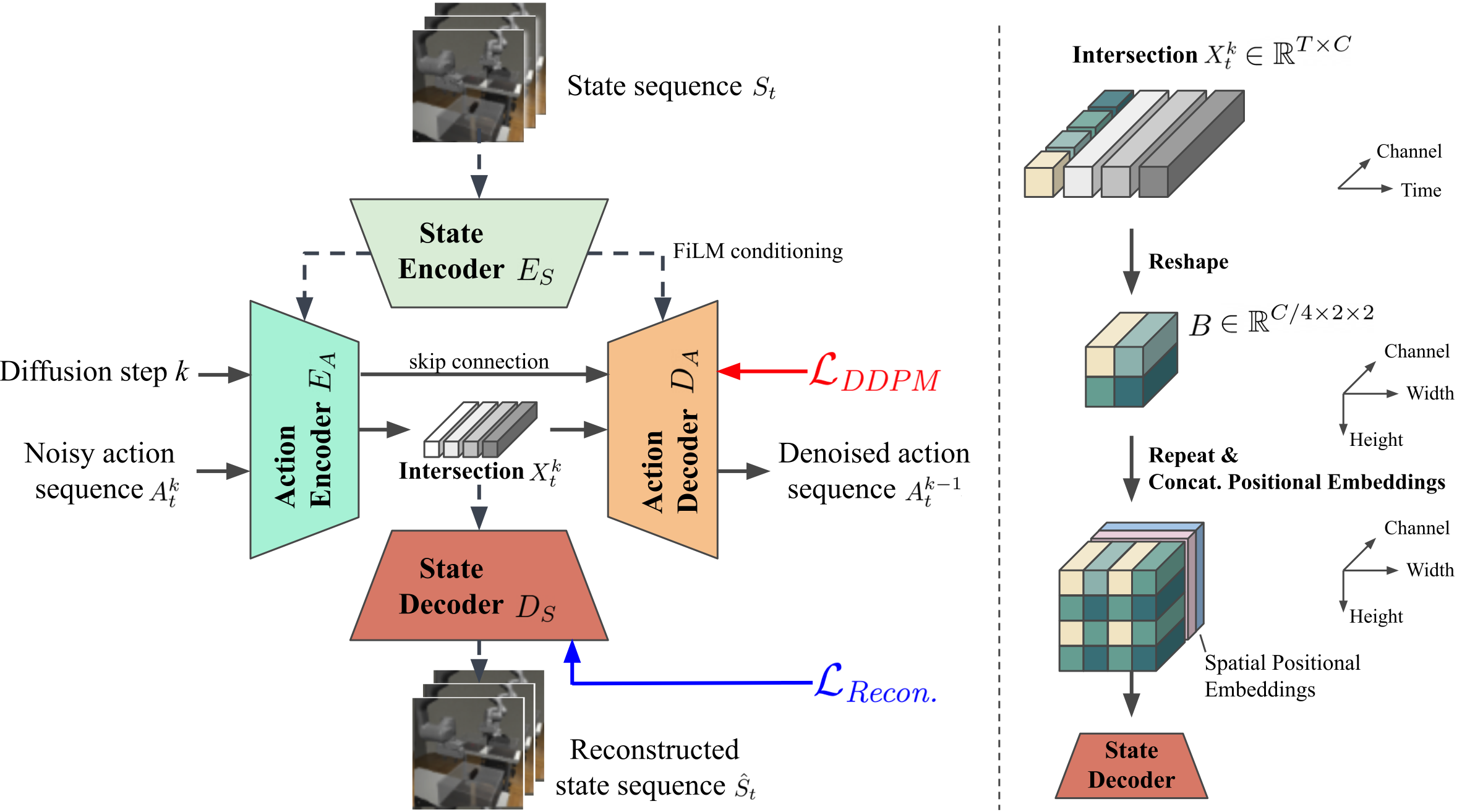}
    \caption{Left: Architecture of Crossway Diffusion. We introduce a state decoder to the existing Diffusion Policy~\cite{chi2023diffusion} as well as an auxiliary reconstruction objective $\mathcal{L}_{\textit{Recon.}}$. The state decoder takes a transformed intermediate representation `intersection' to reconstruct the input states. Right: Transformation applied to `intersection' for the visual state decoder.}
    \label{fig:main_arch}
\end{figure}

Crossway Diffusion extends existing Diffusion Policy~\cite{chi2023diffusion} by introducing 
(1) a state decoder and
(2) an auxiliary objective, both for reconstructing the input states. 
The overall architecture of Crossway Diffusion is presented as Fig.~\ref{fig:main_arch}.
Specifically, the state decoder takes the intermediate representation of the diffusion process $X_t^k$ to reconstruct the input states.
The reconstruction objective is jointly optimized with the diffusion loss $\mathcal{L}_{\textit{DDPM}}$ during training. 

\subsection{State Decoder}~\label{sec:method_state_decoder}
The state decoder $D_S$ is newly introduced to reconstruct the input states from the intermediate representation along with the existing action sequence denoising pipeline.
The new procedure is summarized as $\hat{S}_t=D_S\left( g\left(X_t^k \right) \right)$, where $\hat{S}_t$ is the reconstructed state and $g(\cdot)$ is the \emph{intersection transformation}, discussed in the following section. 

The representation $X_t^k$ is dubbed as \emph{intersection} since both the flow of action denoising and the flow of state reconstruction pass through this tensor and then head to their corresponding destinations.
For the same reason, we name our method \emph{Crossway Diffusion}, whose key feature is to have two flows of information intersected in the diffusion model during the training phase.
By doing so, we explicitly regularize $X_t^k$ to capture the information from both flows, which benefits representation learning. 
Our design includes two distinct types of state decoders: one for visual states and another for low-dimensional states.
For each source of the states (multiple cameras, joint angles, and so on), we assign a dedicated decoder for the best reconstruction results. 

\begin{figure}[tb]
    \centering
    \includegraphics[width=0.485\textwidth]{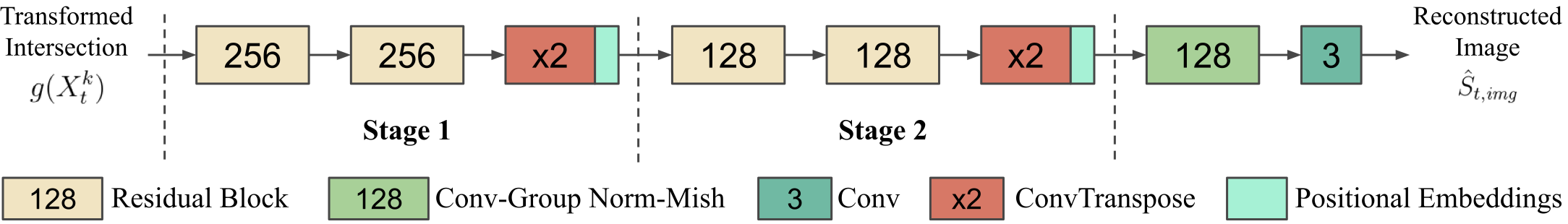}
    \caption{Architecture of visual state decoder. Numbers in the blocks indicate the number of output channels except that ConvTranspose doubles the spatial resolution while keeping the number of channels unchanged.}
    \label{fig:current_dec}
\end{figure}

To reconstruct the visual states, the visual state decoders are made of a sequence of 2D residual convolutional blocks, transposed convolutional layers for upsampling (ConvTranspose), and vanilla convolutional layers (See Fig.~\ref{fig:current_dec}).
The numbers in the blocks indicate the number of output channels except for ConvTranspose.
ConvTranspose doubles the spatial resolution of the input tensor while keeping the number of channels unchanged.
Inspired by \cite{watters2019spatial} and \cite{mildenhall2020nerf}, we adapt the positional embedding from~\cite{vaswani2017attention} to present the pixel coordinates. 
The positional embedding is concatenated to the output of ConvTranspose along the channel axis.

Other low-dimensional states are regressed by three-layer MLPs. 
The widths of the hidden layers are in the ratios of 4:2:1 compared to the width of the low-dimensional states.

Note that the reconstruction with the state decoder $D_S$ is only used during the training,
serving as an `interpreter' to generate additional supervisory signals to train better intermediate representations.
The state decoder will be discarded after the training.



\subsection{Intersection Transformation}

\label{sec:method_trans}
Intersection transformation $g(\cdot)$ adapts the intermediate representation $X_t^k$ for the state decoder to pursue the reconstruction task.
Assume $X_t^k$ is split along the time axis as a list of vectors $X_t^k := \left\{ X_{t, i}^k | i=1, 2, ..., T \right\}$. 
By default, only the first vector in the list $X_{t, 1}^k$ is sent to the state decoder for state reconstructions. 

The transformation for the low-dimensional state decoder $g_{\textit{low-dim}}(\cdot)$ is an identity function which means the state decoder directly takes $X_{t, 1}^k$ as the input.

In contrast, the visual state decoder employs another transformation $g_{\textit{img}}(\cdot)$ due to the significant dimensional disparity between $X_{t, 1}^k$ and the visual states.
As shown in the right of Fig.~\ref{fig:main_arch}, to bridge the dimensionality gap, the $C$ elements of $X_{t, 1}^k$ are equally split into 4 folds and then reshaped as a $ C / 4 \times 2 \times 2$ block $B$.
The block $B$ is then repeated multiple times in two spatial dimensions so that it will have a spatial resolution of a quarter of the desired reconstructed image along each spatial dimension. 
Finally, we concatenate the same positional embeddings introduced in the previous section along the channel axis before sending the transformed intersection to the visual state decoder.

We further explore the effect of multiple designs of $g(\cdot)$ in Section~\ref{sec:ablation}.

\subsection{Crossway Diffusion Loss}
Given the state decoder and the state reconstruction task, 
we introduce a reconstruction loss $\mathcal{L}_{\textit{Recon.}}$ between the reconstructed states and the original input states, which provides an auxiliary training signal to $D_S$, $E_S$, and $E_A$.
$\mathcal{L}_{\textit{Recon.}}$ is simply implemented as a Mean Squared Error (MSE).

In addition to $\mathcal{L}_{\textit{DDPM}}$ used in Diffusion Policy~\cite{chi2023diffusion} (Eq.~\ref{eq:ddpm}), which supervises $D_A$, $E_A$ and $E_S$, 
$\mathcal{L}_{\textit{Recon.}}$ is jointly optimized with $\mathcal{L}_{\textit{DDPM}}$ by a simple weighted addition.
That is, all network modules $E_A$, $D_A$, $E_S$, and $E_S$ are trained jointly.
The total loss for Crossway Diffusion is denoted as Eq.\ref{eq:loss} 
and we find $\alpha=0.1$ is a generally good setting without extensive hyperparameter search.
\begin{equation}
\begin{split}
    \mathcal{L}_{\textit{Recon.}} &= \mathrm{MSE}(S_t, \hat{S}_t)\\
    \mathcal{L}_{\textit{Crossway}} &= \mathcal{L}_{\textit{DDPM}} + \alpha \mathcal{L}_{\textit{Recon.}}
    \label{eq:loss}
\end{split}
\end{equation}

\section{Experiment}
We first evaluate Crossway Diffusion on multiple challenging simulated tasks and two real-world tasks. 
Then, we explore variants of Crossway Diffusion regarding state decoder design and the auxiliary objectives.
Through an extensive amount of experiments, we confirm that Crossway Diffusion consistently leads to better performance than vanilla Diffusion Policy~\cite{chi2023diffusion} and other baselines~\cite{florence2022implicit} over all the tested tasks.

\subsection{Task and Dataset}
We follow~\cite{chi2023diffusion} and choose five tasks Can, Lift, Square, Transport, and Tool~Hang from Robomimic~\cite{mandlekar2021matters} and Push-T from Implicit Behaviour Cloning~(IBC)~\cite{florence2022implicit}.
In addition, we build a real-world robot arm manipulation environment and collect our own data for two tasks.

In the `Can' task, the robot needs to lift a soda can from one box and put it into another box.
In the `Lift' task, the robot needs to lift a cube above a certain height.
In the `Square' task, the robot needs to fit the square nut onto the square peg.
The `Transport' task entails the collaborative effort of two robot arms to transfer a hammer from a closed container on one table to a bin on another table.
One arm is responsible for retrieving and passing the hammer, while the other arm cleans the bin and receives the passed hammer.
In `Tool~Hang', 
the robot needs to insert the hook into the base to assemble a frame and then hang a wrench on the hook.
Additionally, the `Push-T' task involves pushing a T-shaped block (gray) onto a target location (green) in a 2D space.
We investigate both two types of datasets for all tasks if available: `ph' - proficient-human demonstration and `mh' - multi-human demonstrations, originally defined by Robomimic~\cite{mandlekar2021matters}. 
`mh' is designed to have diverse proficiency on the task compared to `ph'.

In the real-world environment, we set up a robot arm with a gripper and two RGB webcams.
The first camera is stationary and offers a third-person perspective of the operating space, while the second camera is mounted on the gripper, providing a first-person view for grasping.
The action space encompasses both the 3D position of the robot arm and a binary signal for the gripper's opening and closing.
For task `Duck~Lift', the objective is to lift a rubber duck above a certain height.
For task `Duck~Collect', the robot needs to collect four ducks and sort them into two separate containers based on the colors of the ducks.

\begin{figure}
    \centering
    \begin{minipage}{0.1\textwidth}
        \includegraphics[width=1\linewidth]{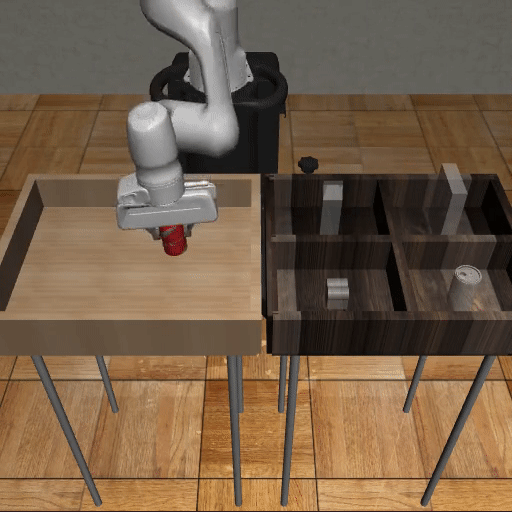}
        \centering{Can}
    \end{minipage}
    \begin{minipage}{0.12\textwidth}
        \includegraphics[width=0.83\linewidth]{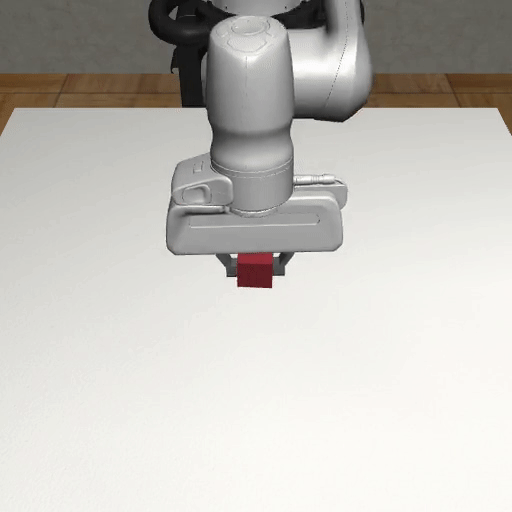}
        \centering{Lift}
    \end{minipage}
    \begin{minipage}{0.1\textwidth}
        \includegraphics[width=1\linewidth]{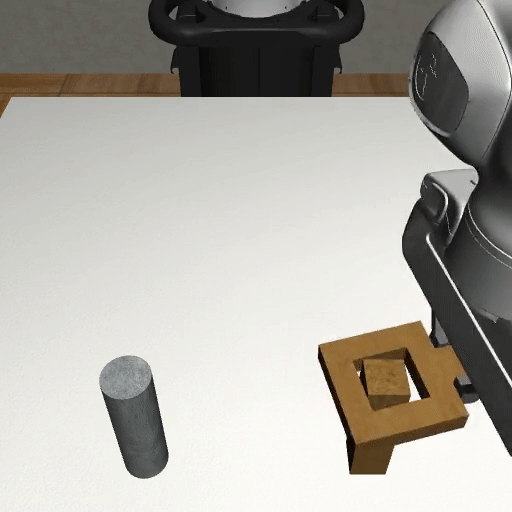}
        \centering{Square}
    \end{minipage}
    \begin{minipage}{0.12\textwidth}
        \includegraphics[width=0.83\linewidth]{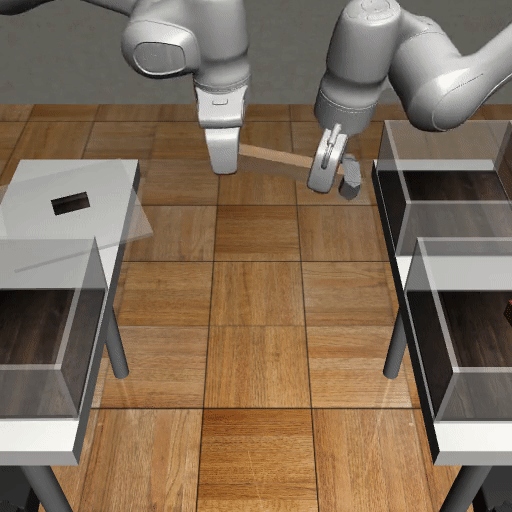}
        \centering{Transport}
    \end{minipage}\\
    \begin{minipage}{0.1\textwidth}
        \includegraphics[width=1\linewidth]{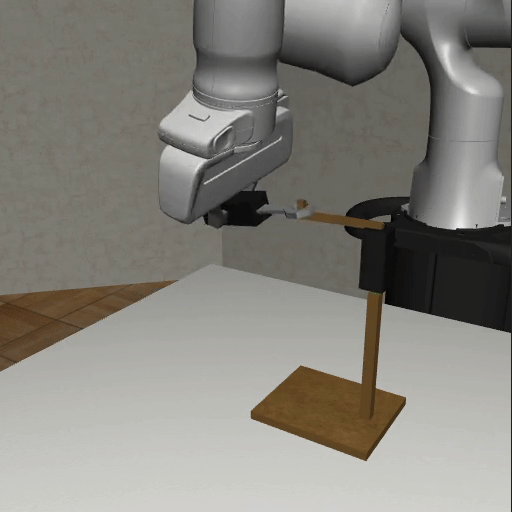}
        \centering{Tool~Hang}
    \end{minipage}
    \begin{minipage}{0.12\textwidth}
        \includegraphics[width=0.83\linewidth]{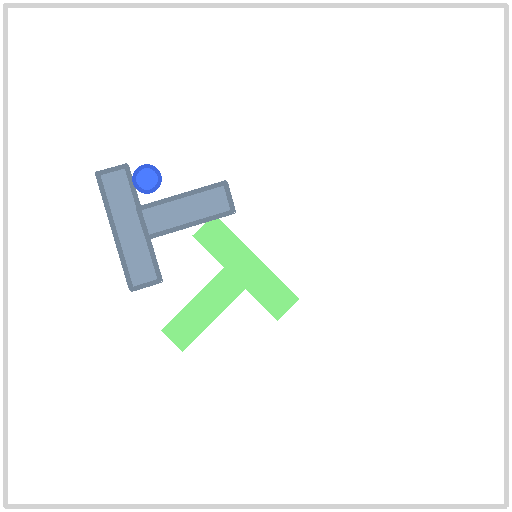}
        \centering{Push-T}
    \end{minipage}
    \begin{minipage}{0.1\textwidth}
        \includegraphics[width=1\linewidth]{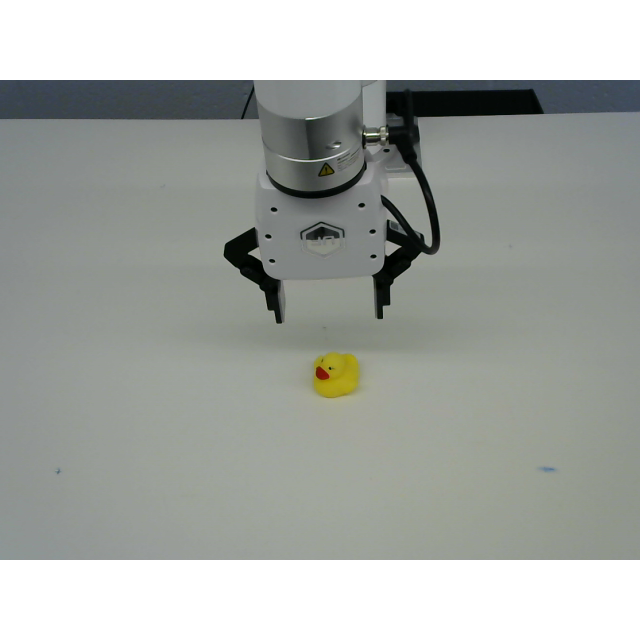}
        \centering{Duck~Lift}
    \end{minipage}
    \begin{minipage}{0.12\textwidth}
        \includegraphics[width=0.83\linewidth]{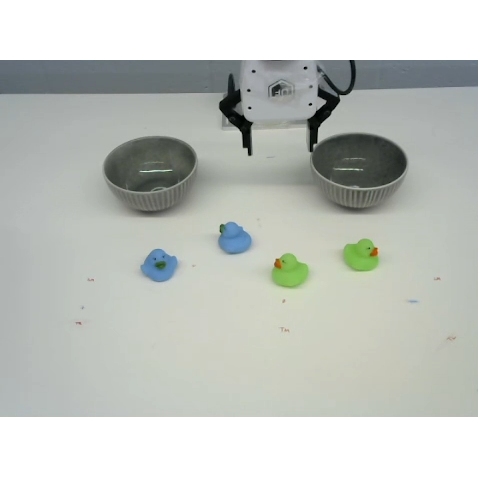}
        \centering{Duck~Collect}
    \end{minipage}
    \caption{Visual reference for all tasks}
    \label{fig:dataset}
\end{figure}

A visual reference for all the tasks is provided as Fig.~\ref{fig:dataset} and an information summary is presented in Tab.~\ref{tab:dataset}.

\begin{table}[t]
  \caption{Dataset summary. ph: the number of proficient-human demonstrations; mh: the number of multi-human demonstrations; R?: whether the dataset is a real-world dataset or not; Rob.: the number of robots; Obj.: the number of objects; Cam.: the number of cameras; Act-D: action dimension;  Steps: max number of rollout steps. }
  \label{tab:dataset}
  \centering
\resizebox{0.45\textwidth}{!}{
  \begin{tabular}{c|ccc|ccccc}
    \toprule
    \textbf{Task} & \textbf{ph}  &  \textbf{mh} &  \textbf{R?} & \textbf{Rob.}  &   \textbf{Obj.} & \textbf{Cam.} &  \textbf{Act-D}    & \textbf{Steps}  \\
    \midrule
    Can & 200 & 300 &  N &1 &  1 & 2   & 7  & 400\\
    Lift & 200 & 300 &  N &1 &  1 & 2   & 7  & 400\\
    Square & 200 & 300 &  N &1 &  1 & 2   & 7  & 400\\
    Transport & 200 & 300 & N & 2 & 3 & 4 & 14 & 700  \\    
    Tool~Hang & 200 & - & N & 1 & 2 & 2 & 7  & 700  \\ 
    Push-T & 200 & - & N & 1 & 1 & 1    & 2  & 300  \\ 
    \midrule
    Duck~Lift & 100 & -  &  Y  & 1 & 1 & 2 & 4  &  50 \\ 
    Duck~Collect & 100 & -  &  Y  & 1 & 1 & 2 & 4  &  200 \\ 
    \bottomrule
  \end{tabular}
}
\end{table}

\subsection{Main Results}
\subsubsection{Evaluation Metrics}
Consistent with prior studies, for the Push-T task, we measure the extent of target location coverage achieved by the T block, which is the ratio of the covered area to the total area.
We adopt the success rate of other tasks as a performance metric.
All the models are trained for $500$ epochs and evaluated at the end of training.
Note that we train and evaluate all methods from scratch with our setting for a fair comparison, as the numbers from \cite{chi2023diffusion} are not reliable because of the bug in their evaluation code\footnote{Please check this link for more details \url{https://github.com/real-stanford/diffusion_policy/issues/6}.}.
For all diffusion-based methods, we benchmark the exponential moving average (EMA) version of the model for better stability, as suggested by~\cite{ho2020denoising}.
More training details are available in Section~\ref{sec:train_detail}.

\subsubsection{Simulated Experiments} 
For all the simulated tasks, in Tab.~\ref{tab:res_sim} we report the average performance over 1000 randomly initialized episodes $\times$ 3 models trained with random seeds, as well as the standard deviation among random seeds.
That is, each score before ± in Tab.~\ref{tab:res_sim} is an average of 3000 episodes, and each number after ± is the standard deviation of the score of 3 random seeds.

\begin{table*}[t]
\caption{Scores on simulated datasets. We report the average of 3000 episodes and the standard deviation of 3 seeds}
\label{tab:res_sim}
\centering
\resizebox{\textwidth}{!}{
\begin{tabular}{ccccccccccc}
\toprule
Method &        Can,~ph &        Can,~mh &       Lift,~ph &       Lift,~mh &     Square,~ph & Square,~mh &  Transport,~ph &  Transport,~mh &      Tool~Hang,~ph &         Push-T\\
\midrule
LSTM-GMM                   &  0.714 ± 0.247 &  0.887 ± 0.033 &  0.978 ± 0.017 &  0.992 ± 0.001 &  0.643 ± 0.023  &    0.491 ± 0.057 &  0.656 ± 0.049 &  0.254 ± 0.017 &    0.460 ± 0.060 &  0.567 ± 0.013 \\
IBC~\cite{florence2022implicit}                 &  0.008 ± 0.006 &  0.001 ± 0.001 &  0.709 ± 0.008 &  0.222 ± 0.112 &  0.002 ± 0.001   &  0.000 ± 0.001 &      0.000 ± 0.000 &      0.000 ± 0.000 &      0.000 ± 0.000 &  0.687 ± 0.031 \\
Diffusion Policy CNN~\cite{chi2023diffusion}       &  0.992 ± 0.002 &  0.958 ± 0.003 &      \textbf{1.000 ± 0.000} &  \textbf{0.998 ± 0.001} &  \textbf{0.935 ± 0.006}  &  0.858 ± 0.007 &  0.859 ± 0.015 &  0.643 ± 0.004 &  0.772 ± 0.012 &  0.819 ± 0.002 \\
\midrule
\textbf{Crossway Diffusion (Ours)} &  \textbf{0.994 ± 0.002} &  \textbf{0.965 ± 0.003} &      \textbf{1.000 ± 0.000} &    \textbf{0.998 ± 0.000} &  \textbf{0.935 ± 0.005} & \textbf{0.879 ± 0.010} &  \textbf{0.864 ± 0.016} &     \textbf{0.800 ± 0.020} &  \textbf{0.792 ± 0.014} &   \textbf{0.843 ± 0.020} \\
\bottomrule
\end{tabular}
}
\end{table*}
From the comparison, the proposed Crossway Diffusion consistently outperforms the baseline Diffusion Policy \cite{chi2023diffusion} in all datasets, as well as other baselines. 
We observe an improvement of $15.7\%$ over the success rate in `Transport,~mh', emphasizing the effectiveness of our method when the demonstration data is varied in proficiency.
Please refer to Section~\ref{sec:example_traj} and ~\ref{sec:diff_viz} for our example episodes and visualization of the action generation process.

\subsubsection{Real-world Experiments}
\begin{figure}[tb]
    \centering
    \includegraphics[width=0.40\textwidth, trim={2.3cm 3cm 1cm 4cm},clip]{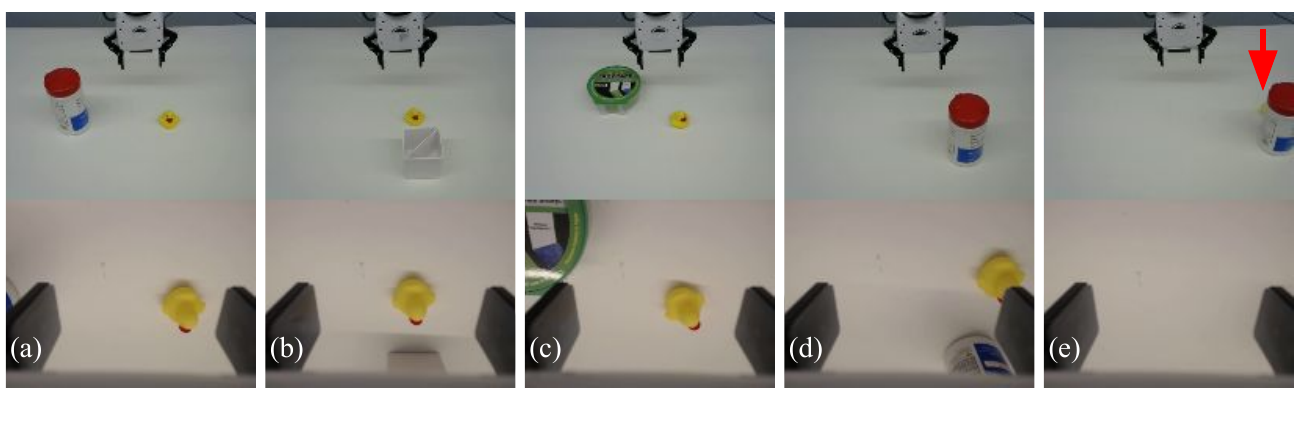}
    \caption{Duck~Lift task under different obstructions. Two rows show the images captured by two cameras respectively. The red arrow in (e) is used to indicate the position of the duck.}
    \label{fig:obstructions}
\end{figure}

For all tested methods, we run `Duck~Lift' for 20 episodes and `Duck~Collect' for 10 episodes and measure the success rate.
Note that in task `Duck~Collect', one episode is considered a success only if the robot successfully collects and sorts all four ducks.
For each episode, we place the ducks at random initial positions but keep the positions consistent across tested methods. 
The results are reported in Tab.~\ref{tab:res_real}, highlighting the advantages of our method.
\begin{table}[t]
    \centering
    \caption{Success rate of real-world tasks}
    \resizebox{0.4\textwidth}{!}{
    \begin{tabular}{ccc}
        \toprule
         & Duck~Lift & Duck~Collect \\
         \midrule
         Diffusion Policy CNN~\cite{chi2023diffusion} & 0.80 & 0.70 \\
         \textbf{Crossway Diffusion (Ours)} & \textbf{0.95} & \textbf{0.80}\\
         \bottomrule
    \end{tabular}
    }
    \label{tab:res_real}
\end{table}
We also show that our method is robust to distractions like unseen objects, occlusions in one camera, and other distractions (Fig.~\ref{fig:obstructions}(a-d)). 
Please refer to Section~\ref{sec:robust} and the video for more information.

Additionally, we provide qualitative results on image reconstruction in Section~\ref{sec:more_recon}, where the quality of reconstructed visual states is surprisingly high.

\subsection{Ablations}
\label{sec:ablation}
The ablation study mainly focuses on two critical designs in our method: the state decoder and the auxiliary objective. 

\subsubsection{On State Decoder}
\label{sec:ablation_dec}
Experiments in this section are designed to answer (1) which representation for state reconstruction benefits policy learning most, as well as (2) what is the best architecture for the visual state decoder.

\begin{figure}[tb]
    \centering
    \includegraphics[width=0.48\textwidth]{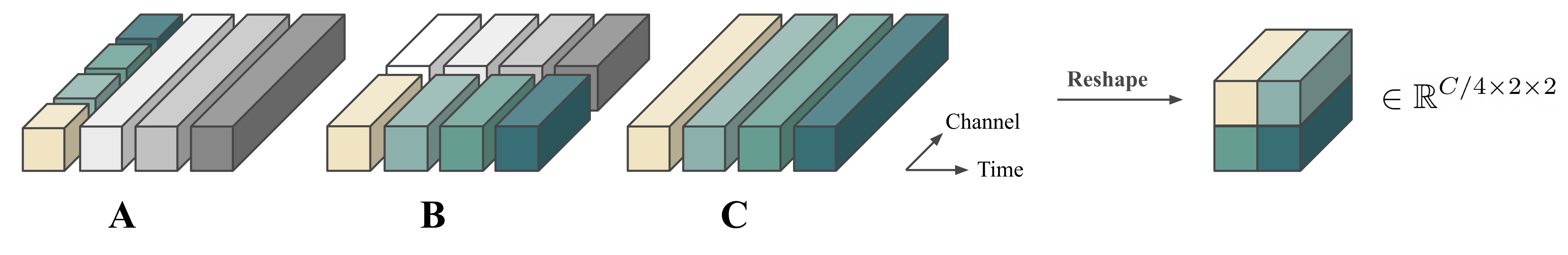}
    \caption{Selected parts for state reconstruction in three designs. White and gray parts are not used for reconstruction.}
    \label{fig:ablation}
\end{figure}

First, two more designs (Design B and C) of the intersection transformation presented in Fig.~\ref{fig:ablation} are studied over five simulated datasets, as well as one variant (Design D) that utilizes $h_t$ instead of intersection $X_t^k$ for reconstruction. 

Design A is the default Crossway Diffusion introduced in Section~\ref{sec:method}.
In Design B, the first $C / 2$ channels for all vectors in intersection $X_t^k$ are selected for reconstruction, while the rest is left dedicated to the denoising process.
In contrast, Design C takes advantage of all vectors in $X_t^k$ for the reconstruction.
Additionally, for both Design B and C, the selected vectors are independently projected by a linear layer to match the target number of channels $C / 4$.
The latter operations like reshape and repeat in Fig.~\ref{fig:main_arch} are kept the same.
Finally, in Design D, the state decoder directly takes the output of the state encoder $h_t$ instead of $X_t^k$, which disentangles the action denoising flow and the state reconstruction flow.
A linear layer is also applied to project $h_t$ to the target dimension $C / 4$ in the channel axis.
\begin{table}[t]
\caption{Ablations on intersection transformation.}
\label{tab:ablation_trans}
\centering
\resizebox{0.49\textwidth}{!}{
\begin{tabular}{ccccccc}
\toprule
 &     Square,~mh &  Transport,~ph &  Transport,~mh &      Tool~Hang,~ph &         Push-T \\
\midrule

A (default) &   0.879 ± 0.010 &  0.864 ± 0.016 &     0.800 ± 0.020 &  \textbf{0.792 ± 0.014} &   \textbf{0.843 ± 0.020} \\
B &  \textbf{0.881 ± 0.017} &   0.882 ± 0.010 &  0.784 ± 0.025 &   0.777 ± 0.010 &  0.835 ± 0.012 \\
C &  0.868 ± 0.006 &  \textbf{0.906 ± 0.012} &  \textbf{0.814 ± 0.028} &  0.783 ± 0.005 &  0.831 ± 0.003 \\
\midrule
D &  0.873 ± 0.012 &  0.892 ± 0.002 &  0.764 ± 0.013 &   0.790 ± 0.007 &  0.819 ± 0.015 \\
\midrule
\graytext{Diff.~\cite{chi2023diffusion}}        &  \graytext{0.858 ± 0.007} &  \graytext{0.859 ± 0.015} &  \graytext{0.643 ± 0.004} &  \graytext{0.772 ± 0.012} &  \graytext{0.819 ± 0.002} \\
\bottomrule
\end{tabular}
}
\end{table}

The results presented in Tab.~\ref{tab:ablation_trans} show that Design A, B, and C consistently outperform the baseline Diffusion Policy CNN~\cite{chi2023diffusion}~(Diff. in table).
However, Design D achieves similar results to the baseline on Push-T and shows improvements on other datasets.
Such observations validate the importance of not only the \emph{reconstruction task} but also the design which \emph{explicitly forces two flows to intersect with each other}.
Though these designs show diverse advantages over different tasks, we choose Design A as the default due to its computational simplicity.

Then, the variants with different visual state decoder architectures are studied.
The default visual state decoder contains two stages where each stage has two residual blocks with a transposed convolution for spatial upsampling  (see Fig.~\ref{fig:current_dec}).
\emph{Shallower Dec.} takes the second stage out and only keeps the first stage. 
For \emph{ViT Dec.}, the visual state decoder is a two-layer Vision Transformer (ViT)~\cite{dosovitskiy2020image} whose token dimension is $C/4$ and the positional embeddings are directly added to the tokens before the ViT instead of concatenation.

The results on Push-T are presented in Tab.~\ref{tab:ablation_dec} (more available in Appendix~\ref{sec:app_ablation}), showing that all variants of our methods perform better than the baseline Diffusion Policy CNN~\cite{chi2023diffusion}~(Diff. in table).
Such observation validates the effectiveness of the reconstruction task and our specific visual state decoder design.
\begin{table}[b]
    \centering
    \caption{Ablation results tested on Push-T}
    \label{tab:ablation_dec}
    \resizebox{0.49\textwidth}{!}{
    \begin{tabular}{c|cc|c|c}
    \toprule
     Default &   Shallower Dec. &  ViT Dec. &  Visual-only & \graytext{Diff.~\cite{chi2023diffusion}} \\
    \midrule
      \textbf{0.843 ± 0.020} &  0.822 ± 0.014 &  0.824 ± 0.008 &  0.828 ± 0.012 &      \graytext{0.819 ± 0.002} \\
    \bottomrule
    \end{tabular}
    }
\end{table}

\subsubsection{On auxiliary objective}
\label{sec:ablation_ssl}
By default, both visual states and low-dimensional states are reconstructed in Crossway Diffusion.
We first benchmark a simple variant called \emph{Visual-only}, which reconstructs only the visual states.
Results in Tab.~\ref{tab:ablation_dec} verify the benefit of predicting all the input states.

Then we test the variants where the state decoder predicts the state that is $N$ steps ahead of the current state, instead of predicting (reconstructing) the current state as in the default setting ($N=0$).
From Tab.~\ref{tab:ablation_predict} we empirically learn that reconstructing the current state is the most beneficial method while predicting future states may compromise the performance, even worse than the baseline.
We further extend the study to multiple challenging datasets and obtain a similar observation included in Appendix~\ref{sec:app_ablation}.

\begin{table}[tb]
    \centering
    \caption{Ablations on future prediction, tested on Push-T}
    \label{tab:ablation_predict}
    \resizebox{0.49\textwidth}{!}{
    \begin{tabular}{ccccc}
    \toprule
     $N=0$ (default) &            $N=2$ &            $N=4$ &            $N=6$ &            $N=8$ \\
    \midrule
     \textbf{0.843 ± 0.020} &  0.818 ± 0.006 &  0.827 ± 0.014 &  0.817 ± 0.003 &  0.803 ± 0.013 \\
    \bottomrule
    \end{tabular}

    }
\end{table}

Furthermore, we verify the effectiveness of the proposed reconstruction task in comparison to another SSL objective using contrastive learning inspired by CURL~\cite{laskin2020curl}.
In particular, we independently perform random crop on all images of an observation sequence $S_{t}$ twice to get two augmented sequences $S_{t,a1}$ and $S_{t,a2}$.
The model takes $S_{t,a1}$ to produce intersection $X_{t,a1}^k$ while $S_{t,a2}$ is processed by the exponential moving average (EMA) version of the model to get $X_{t,a2,\mathrm{ema}}^k$.
The intuition is that due to the semantic similarity between $S_{t,a1}$ and $S_{t,a2}$, the intersection $X_{t,a1}$ and $X_{t,a2,\mathrm{ema}}$ should also be similar in a latent space.

We follow CURL~\cite{laskin2020curl} and maintain a learnable matrix $W$. 
For each batch of $b$ samples, we calculate the similarity matrix $M_{\mathrm{sim}}$ between all samples in the same batch using matrix multiplication (first line in Eq.\ref{eq:cl}, where $\mathrm{sg}\left(\cdot \right)$ means stop gradients).
Then the contrastive loss $\mathcal{L}_{\textit{CURL}}$ is formulated as Eq.~\ref{eq:cl}, where $b$ is the batch size and $\alpha=0.1$.
Finally $\mathcal{L}_{\textit{CURL}}$ is jointly optimized with the diffusion loss $\mathcal{L}_{\textit{DDPM}}$ similar to Eq.\ref{eq:loss}.
We name such configuration as \emph{Crossway-CURL}.
\begin{equation}
\begin{split}
    M_{\mathrm{sim}} &= X_{t,a1}^k \cdot W \cdot \mathrm{sg}\left(X_{t,a2,\mathrm{ema}}^k\right)^T\\
    \mathcal{L}_{\textit{CURL}} &= \mathrm{CrossEntropyLoss}\left(M_{\mathrm{sim}}, \mathrm{range}(b) \right)\\
    \mathcal{L}_{\textit{Crossway-CL}} &= \mathcal{L}_{\textit{DDPM}} + \alpha \mathcal{L}_{\textit{CURL}}
\label{eq:cl}
\end{split}
\end{equation}

From Tab.~\ref{tab:ablation_curl}, the contrastive learning variant Crossway-CURL yields much worse performance compared to the baseline on multiple tasks, indicating that not all auxiliary SSL losses benefit policy learning.
Such observations happen to align with the online reinforcement learning case~\cite{li2022does}.
\begin{table}[tb]
    \centering
    \caption{Performance of Crossway-CURL, which adopts a contrastive loss as the auxiliary objective}
    \label{tab:ablation_curl}
    \resizebox{0.49\textwidth}{!}{
    \begin{tabular}{cccccc}
\toprule
 &       Lift,~mh &       Lift,~ph &     Square,~mh &     Square,~ph &         Push-T \\
\midrule
Crossway-CURL  &  0.802 ± 0.024 &  0.678 ± 0.188 &  0.053 ± 0.025 &  0.035 ± 0.007 &   0.518 ± 0.160 \\
Default &    \textbf{0.998 ± 0.000} &      \textbf{1.000 ± 0.000} &   \textbf{0.879 ± 0.010} &  \textbf{0.935 ± 0.005} &   \textbf{0.843 ± 0.020} \\
\midrule
\graytext{Diff.~\cite{chi2023diffusion}}       &  \textbf{\graytext{0.998 ± 0.001}} &      \textbf{\graytext{1.000 ± 0.000}} &  \graytext{0.858 ± 0.007} &  \graytext{0.935 ± 0.006} &  \graytext{0.819 ± 0.002} \\
\bottomrule
\end{tabular}
}
\end{table}





\section{Related Work}
\subsection{Behavioral Cloning}
Behavioral Cloning (BC)~\cite{Pomerleau1988ALVINNAA,kaelbling1993learning,kormushev2011imitation} is a straightforward but surprisingly effective way to obtain robot policies. 
With pre-collected state-action pairs, BC learns a policy like fitting a dataset~\cite{atkeson1997robot}, with additional techniques like reward labeling / Inverse Reinforcement Learning (IRL)~\cite{ng2000algorithms,osa2018algorithmic}, distribution matching~\cite{ho2016generative,peng2021amp}, and incorporating extra information~\cite{florence2019self,zhang2018deep,rahmatizadeh2018vision}. 
Apart from explicitly generating output actions, BC can be done implicitly, where an energy-based model is learned to model the action distribution~\cite{florence2022implicit}. 
Implicit BC is found to be effective in real-world robot tasks. 
BC also boosts some online RL algorithms like TD3+BC~\cite{fujimoto2021minimalist}, DeepMimic~\cite{peng2018deepmimic}, and more~\cite{rajeswaran2017learning}. 
Recent Diffusion-based BC is more like an advanced approach for matching the behavior distribution, which potentially helps mitigate the distribution shift problem in BC~\cite{abbeel2004apprenticeship}.

\subsection{Policy Learning as Sequential Modeling}
Sequential modeling~\cite{dt, trajectorytransformer, starformer} is the recent direction to solve offline-RL or Imitation Learning problems. 
The key is to optimize a policy on a trajectory basis from pre-collected experiences, with a reward signal (offline-RL) or without it (imitation learning). 
To model the trajectory composed by state-action-reward tuples, Transformer~\cite{att} is firstly adopted to this problem in the light of success in modeling natural language. 
In this formulation, state-action-reward tuples are regarded as equal units~\cite{dt, trajectorytransformer} or with Markovian properties~\cite{starformer,hu2023graph} for better long-term modeling. 
There are works to extend the formulation to online learning~\cite{zheng2022online}, hindsight matching~\cite{furuta2022distributional}, and bootstrapping~\cite{bootstrappedtransformer}.
Recently, diffusion models have also been adopted to this problem~\cite{janner2022planning,pearce2023imitating,chi2023diffusion}, on which \cite{zhu2023diffusion} conducts a comprehensive survey. 

\subsection{Self-supervised Learning}
Self-supervised learning (SSL)~\cite{vincent2008extracting} is used to learn data representations without task labels. SSL is commonly used to (pre-)train task-agnostic foundation models~\cite{he2022masked,tong2022videomae,wang2023videomae, brown2020language, chen2020simple}, or is used as an auxiliary task together with other learning paradigms~\cite{li2022does, das2024limited}. Similarly, SSL has multiple ways to combine with policy learning, and we briefly categorize them into two: pre-training with SSL~\cite{ha2018world, sermanet2018time, zhan2020framework, shah2021rrl, stooke2021decoupling, shang2021self, wang2022vrl3, xiao2022masked}, and policy learning jointly with auxiliary SSL tasks~\cite{oord2018representation, igl2018deep, hafner2019learning, yingjun2019learning, yarats2019improving, lee2020stochastic, laskin2020curl, zhu2020masked, mazoure2020deep, lee2020predictive, guo2020bootstrap, schwarzer2020data, zhang2020learning, yu2021playvirtual,kulkarni2019unsupervised,wang2021unsupervised,li2022does}. Studies~\cite{li2022does} have shown that different ways to combine policy learning with SSL have different outcomes. In this work, we follow the joint learning style that optimizes diffusion and reconstruction objectives together.

\section{Conclusion}
In this paper, we investigate how SSL can be used to improve diffusion-based visual behavioral cloning. 
We propose Crossway Diffusion, which involves an extra state decoder and a reconstruction auxiliary objective in addition to the existing diffusion objective during training.
Compared to the baseline, Crossway Diffusion shows consistent and substantial improvements over multiple challenging tasks including two real-world tasks without additional computation during evaluation.
We hope our work inspires further exploration of how to take advantage of the rapidly evolving SSL techniques for better diffusion-based policies.






\clearpage
\bibliographystyle{IEEEtran}
\bibliography{ref}

\clearpage
\section*{Appendix}
\subsection{Training Details}
\label{sec:train_detail}

We primarily adopt the hyperparameters from the Diffusion Policy~\cite{chi2023diffusion}. 
Specifically, the observation horizon $T_s$ and action horizon $T_a$ are set at 2 and 8, respectively.
The learning rate and weight decay are configured at 1e-4 and 1e-6, with a uniform batch size of 64 across all datasets.
Our approach involves 100 diffusion iterations for both the training and inference phases.
Additionally, we explore the impact of reducing denoising steps during inference through DDIM~\cite{ddim}, as elaborated in Appendix~\ref{sec:limitation}. Comprehensive details on image reconstruction hyperparameters and the count of learnable parameters are presented in Tab.~\ref{tab:hp}.


\begin{table}[b]
  \caption{Hyperparameters and the number of learnable parameters. ImgRes: the resolution of visual states (Camera views $\times$ W $\times$ H); CropRes: the image resolution after random crop; RecRes: the resolution of reconstruction target; \#D: number of parameters in diffusion network; \#VE: number of parameters in state encoder; \#VD: number of parameters in state decoder.}
  \label{tab:hp}
  \centering
  \resizebox{0.49\textwidth}{!}{\begin{tabular}{c|cccccc}
    \toprule
    \textbf{Task}   &  \textbf{ImgRes}     
    & \textbf{CropRes}   & \textbf{RecRes}      &  \textbf{\#D}  & \textbf{\#VE}   &  \textbf{\#VD}  \\
    
    \midrule
    Can       &  2$\times$84$\times$84  
    & 2$\times$76$\times$76 & 2$\times$84$\times$84  &    256M                &  22M                   &   6M      \\
    Lift       &  2$\times$84$\times$84  
    & 2$\times$76$\times$76 & 2$\times$84$\times$84  &    256M                &  22M                   &   6M      \\
    Square       &  2$\times$84$\times$84  
    & 2$\times$76$\times$76 & 2$\times$84$\times$84  &    256M                &  22M                   &   6M      \\
    
    Transport       &  4$\times$84$\times$84  
    & 4$\times$76$\times$76 & 4$\times$60$\times$60  &    264M                &  45M                   &   12M      \\
    
    Tool~Hang       &  2$\times$240$\times$240 
    & 2$\times$216$\times$216 & 2$\times$80$\times$80 &    256M                &  22M                   &   6M     \\
    
    Push-T       &  1$\times$96$\times$96  
    & 1$\times$84$\times$84  & 1$\times$96$\times$96  &    252M                & 11M                   &   2M    \\
    \midrule
    Duck~Lift       &  2$\times$160$\times$120   
    & 2$\times$144$\times$108 & 2$\times$80$\times$60 &    255M                 & 22M                    &   6M     \\
    Duck~Collect       &  2$\times$160$\times$120   
    & 2$\times$144$\times$108 & 2$\times$80$\times$60 &    255M                 & 22M                    &   6M     \\
    \bottomrule
  \end{tabular}}
\end{table}

\subsection{Ablations on More Datasets}
\label{sec:app_ablation}
\begin{figure}[b]
    \centering
    \includegraphics[width=0.49\textwidth]{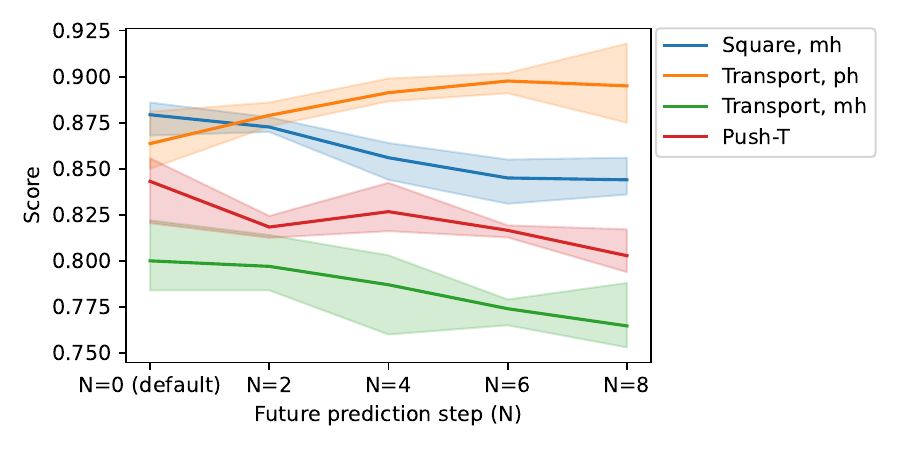}
    \caption{The trend between the score and the number of future prediction steps ($N$).}
    \label{fig:predict_line}
\end{figure}

We extend the ablation experiments for methods introduced in Section~\ref{sec:ablation} to additional datasets. 
 The outcomes related to the state decoder and the auxiliary objectives are respectively tabulated in Tab.~\ref{tab:ablation_dec} and Tab.~\ref{tab:ablation_ssl}.
These results validate the superiority of our default configuration across all datasets, with the exception of `Transport,~ph'.
Interestingly, for the group of experiments where the reconstruction goal is the future states instead of the current state (described in Section~\ref{sec:ablation_ssl}), we empirically find a correlation between the future prediction step ($N$) and the score (shown in Fig.~\ref{fig:predict_line}).
It indicates a decline in performance scores as the number of future prediction steps ($N$) increases. 
However, for `Transport,ph', the optimal performance is achieved at $N = 6$, as depicted in Fig.~\ref{fig:predict_line}.

\begin{table}[b]
    \centering
    \caption{More ablations on the state decoder}
    \label{tab:ablation_state_dec}
    \resizebox{0.49\textwidth}{!}{
    \begin{tabular}{c|cccc}
    \toprule
     & Square,~mh &  Transport,~ph &  Transport,~mh &         Push-T\\
    \midrule
    Ours (default) &   \textbf{0.879 ± 0.010} &  0.864 ± 0.016 &     \textbf{0.800 ± 0.020} &   \textbf{0.843 ± 0.020} \\
    \midrule
    Shallower Dec.     &   0.840 ± 0.021 &  \textbf{0.883 ± 0.002} &   0.699 ± 0.020 &  0.822 ± 0.014 \\
     ViT Dec.                  &  0.859 ± 0.014 &  0.865 ± 0.011 &  0.745 ± 0.025 &  0.824 ± 0.008 \\
    \midrule
    \graytext{Diff.~\cite{chi2023diffusion}}         &  \graytext{0.858 ± 0.007} &  \graytext{0.859 ± 0.015} &  \graytext{0.643 ± 0.004} &  \graytext{0.819 ± 0.002} \\
    \bottomrule
    \end{tabular}
    }
\end{table}

\begin{table}[b]
    \centering
    \caption{More ablations on auxiliary objectives}
    \label{tab:ablation_ssl}
    \resizebox{0.49\textwidth}{!}{
    \begin{tabular}{c|cccc}
    \toprule
     & Square,~mh &  Transport,~ph &  Transport,~mh &         Push-T\\
    \midrule
    Ours (N = 0, default) &   \textbf{0.879 ± 0.010} &  0.864 ± 0.016 &     \textbf{0.800 ± 0.020} &   \textbf{0.843 ± 0.020} \\
    N = 2           &  0.873 ± 0.005 &  0.879 ± 0.007 &  0.797 ± 0.015 &  0.818 ± 0.006 \\
    N = 4           &  0.856 ± 0.011 &  0.891 ± 0.007 &  0.787 ± 0.024 &  0.827 ± 0.014 \\
    N = 6           &  0.845 ± 0.012 &  \textbf{0.898 ± 0.006} &  0.774 ± 0.008 &  0.817 ± 0.003 \\
    N = 8           &  0.844 ± 0.011 &  0.895 ± 0.022 &   0.765 ± 0.020 &  0.803 ± 0.013 \\
    \midrule
    Visual-only              &  0.878 ± 0.001 &  0.869 ± 0.027 &  0.779 ± 0.018 &  0.828 ± 0.012 \\
    \midrule
    \graytext{Diff.~\cite{chi2023diffusion}}         &  \graytext{0.858 ± 0.007} &  \graytext{0.859 ± 0.015} &  \graytext{0.643 ± 0.004} &  \graytext{0.819 ± 0.002} \\
    \bottomrule
    \end{tabular}
    }
\end{table}

\subsection{Discussion on Improving Inference Speed}
\label{sec:limitation}
Even though our self-supervised learning objective improves the performance, it remains the same need for a large number of diffusion iterations during inference as the baseline Diffusion Policy~\cite{chi2023diffusion}. This is still a common challenge for diffusion-based policies that hinder responsiveness in some high-dynamic real-world environments.

\begin{figure}[thb]
    \centering
    \includegraphics[width=0.49\textwidth]{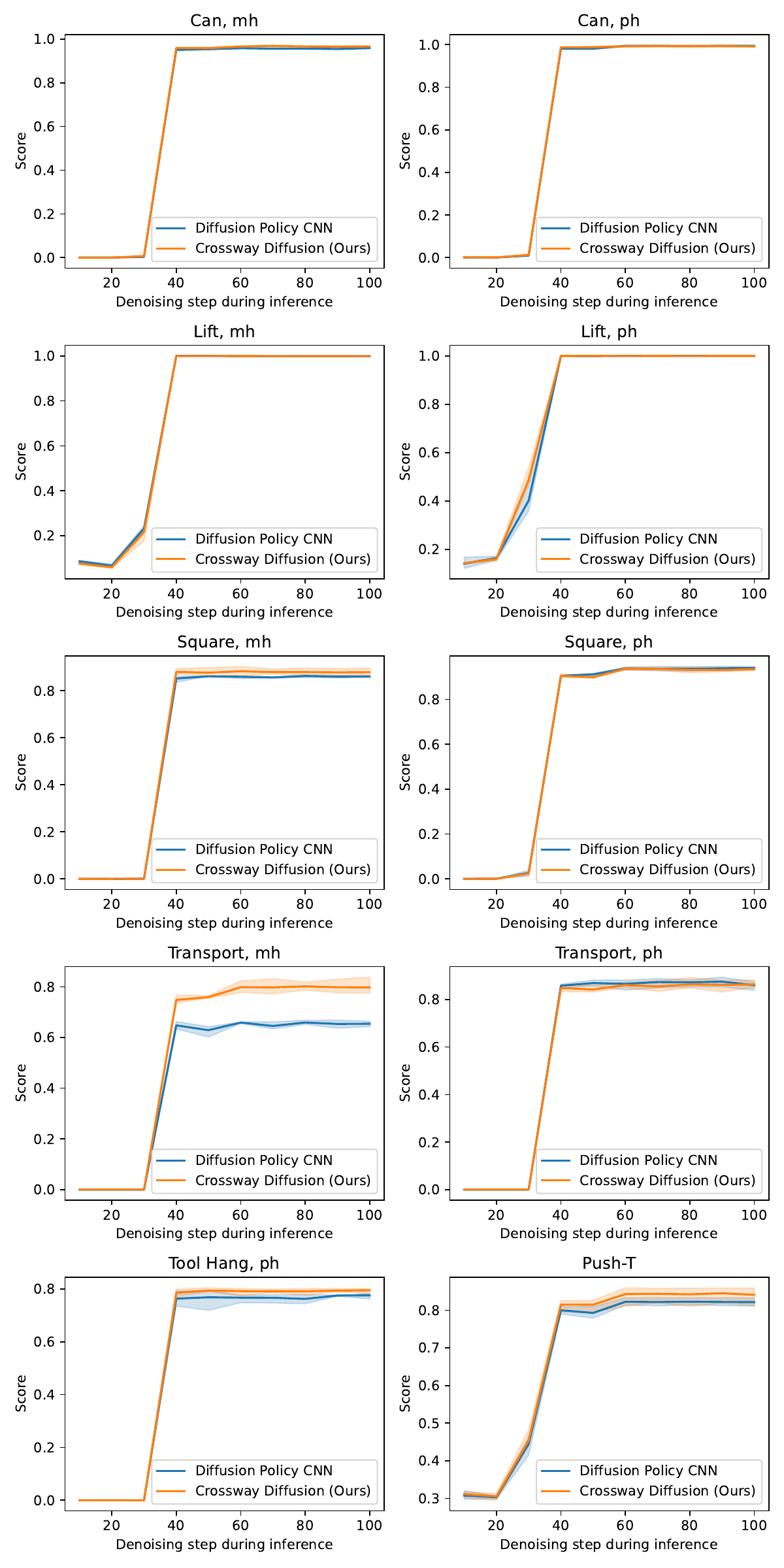}
    \caption{The relationship between the score and the diffusion denoising steps during inference using DDIM~\cite{ddim}.}
    \label{fig:ddim}
\end{figure}

To this end, we use DDIM~\cite{ddim} to accelerate the inference process as suggested in \cite{chi2023diffusion}.
Specifically, we evaluate each trained model by varying the number of diffusion iterations from 10 to 100, in increments of 10.
Fig.~\ref{fig:ddim} demonstrates the relationship between the score and the number of denoising steps during inference.
For each subfigure in ~\ref{fig:ddim}, given three models per method trained using different random seeds, the only difference is the number of denoising steps during inference shown as the x-axis.
In general, the quality of the generated action sequence significantly drops when the number of diffusion iterations gets lower than 40.
However, the score of some challenging tasks starts to drop when the number of diffusion iterations is smaller than 60.

We are actively looking into diffusion acceleration techniques like Progressive Distillation~\cite{salimans2022progressive} and Consistency Models~\cite{song2023consistency} to address the limitation.

\subsection{Visualization of Action Generation Process}
\label{sec:diff_viz}
\begin{landscape}
    \begin{figure}[htb]
    \centering
    \includegraphics[width=1.3\textwidth]{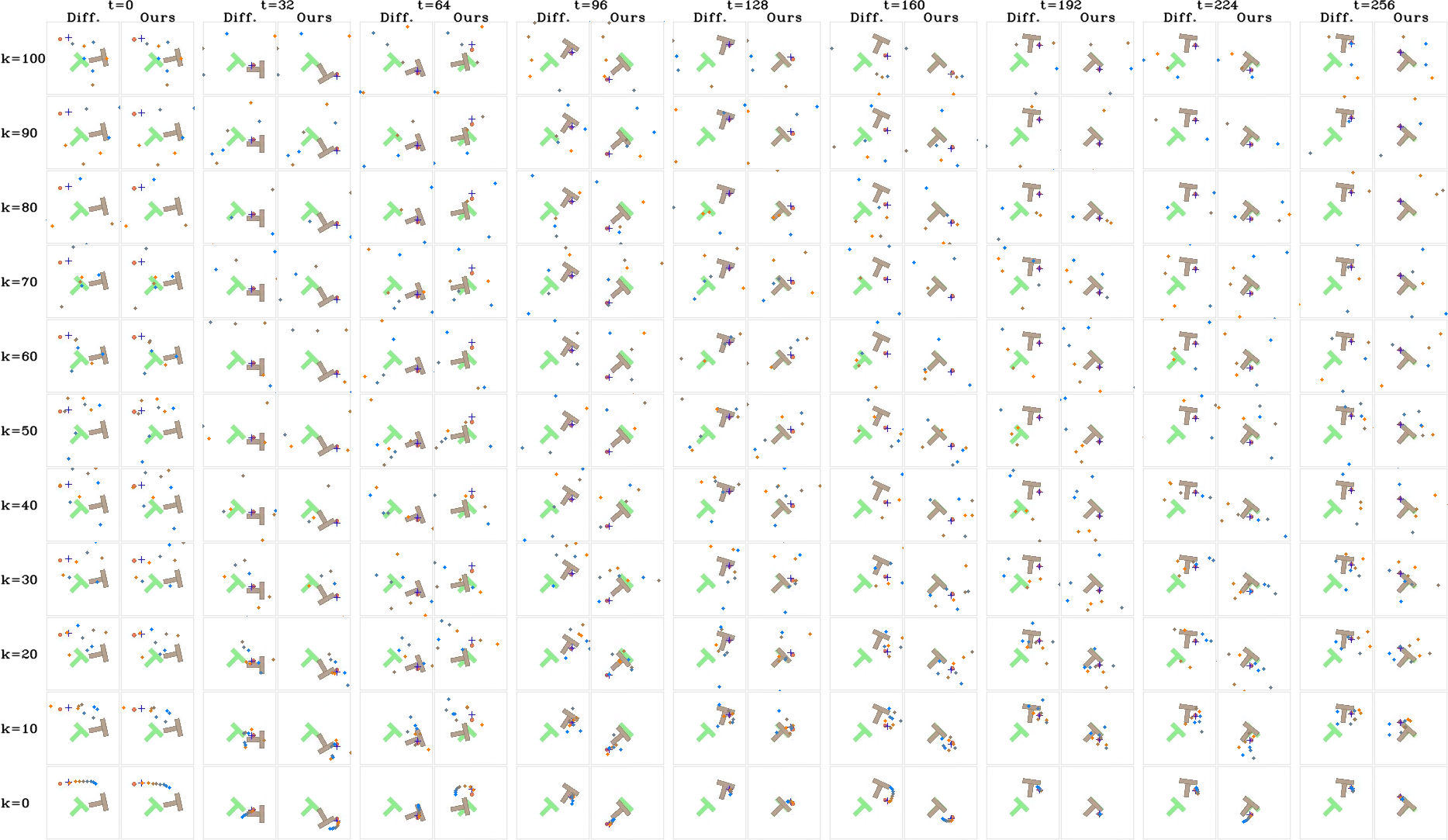}
    \caption{Visualization of the action generation process. $k$ is the diffusion step and $t$ is the state transition step. The first action in the sequence is presented in orange and the last action is in blue. We compare our method with Diffusion Policy~\cite{chi2023diffusion} (Noted as Diff.)} 
    \label{fig:hist_pusht}
    \end{figure}
\end{landscape}

In Fig.~\ref{fig:hist_pusht}, we demonstrate an episode with the action generation (denoising) process for Push-T. 
In the grid, the state transition step $t$ increases from left to right, while the diffusion step $k$ (in other words, noise level) decreases from top to bottom. 
The orange dot stands for the first action in the sequence and the blue dot stands for the last action. 
The colors of the actions in between are linear interpolations of the first and the last colors.
Diffusion Policy~\cite{chi2023diffusion} is also listed for comparison as \emph{Diff.} in the figure.
We strongly recommend readers check the videos at \url{https://youtu.be/9deKHueZBuk} for a better demonstration of Fig.~\ref{fig:hist_pusht} and more real-world robot experiments.

\subsection{Qualitative Test of Real-world Robustness}
\label{sec:robust}

This section presents the qualitative test results of the robustness of our method under various obstructions.

The examples of `Duck~Lift' are shown in Fig.~\ref{fig:obstructions}. 
Fig.~\ref{fig:obstructions}~(a)(b)(c) contain different unseen objects placed on the table as a distraction. 
None of these obstructions are present in the training dataset. 
In Fig.~\ref{fig:obstructions}~(d), the duck is only visible to the second camera. 
Our method successfully executes the task in all the above scenarios indicating robustness to obstructions. 
Fig.~\ref{fig:obstructions}~(e) shows a scenario where the duck is not clearly visible to both cameras. 
In this case, as expected our method fails to locate and lift the duck.

\begin{figure}[tb]
    \centering
    \includegraphics[width=0.49\textwidth]{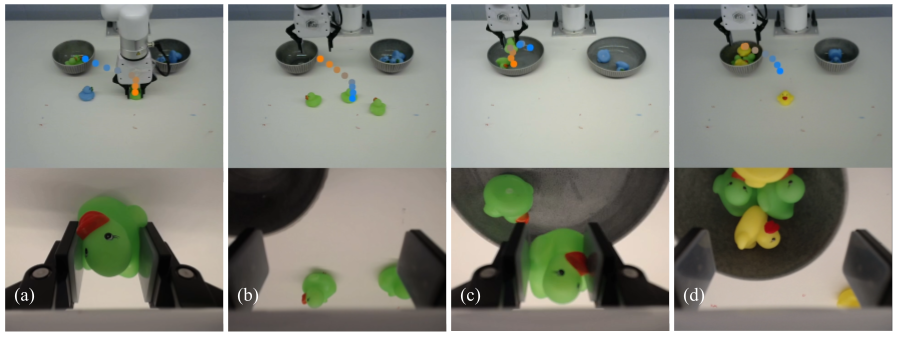}
    \caption{Duck~Collect task under different obstructions. Two rows show the images captured by two cameras respectively. (a)(b): Additional ducks. (c): Unseen containers. (d): Unseen duck color. The dots are added to the first row to demonstrate the trajectories generated from the model. The orange dot notes the beginning of an action sequence while the blue dot notes the end.}
    \label{fig:collect_obs}
\end{figure}

The examples of `Duck~Collect' are shown in Fig.~\ref{fig:collect_obs}.
Fig.~\ref{fig:collect_obs}~(a)(b) contain additional ducks beyond the four in the training set.
In Fig.~\ref{fig:collect_obs}~(c), we replace the containers with unseen ones.
Our method successfully executes the task in all the above scenarios, highlighting the robustness of our method. 
In Fig.~\ref{fig:collect_obs}~(d), we bring the ducks with an unseen yellow color to the agent, and our method can still pick up the ducks and put them consistently into the same container.

\subsection{Qualitative Results on Image Reconstruction}
\label{sec:more_recon}
\begin{figure*}
    \centering
    \begin{minipage}{0.12\textwidth}
        \includegraphics[width=\linewidth]{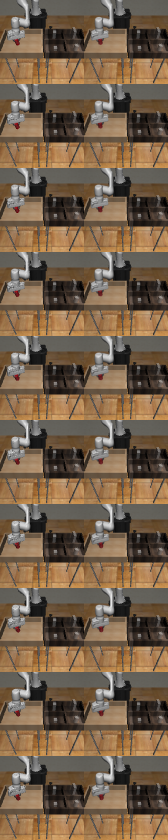}
        \includegraphics[width=\linewidth]{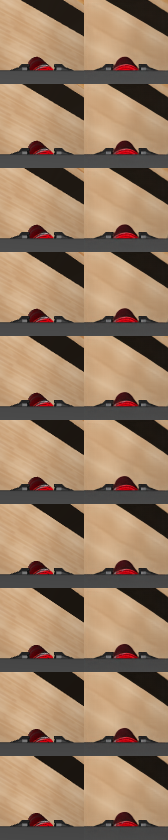}
        \centering{Can,~mh}
    \end{minipage}
    \begin{minipage}{0.12\textwidth}
        \includegraphics[width=\linewidth]{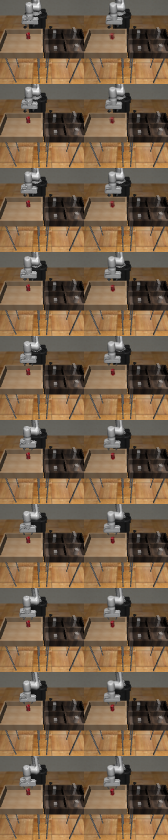}
        \includegraphics[width=\linewidth]{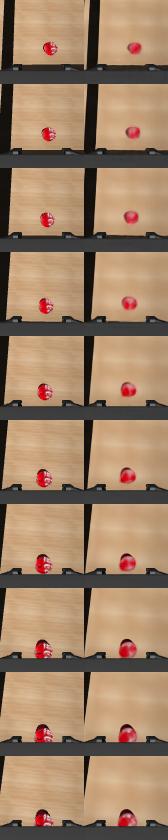}
        \centering{Can,~ph}
    \end{minipage}
    \begin{minipage}{0.12\textwidth}
        \includegraphics[width=\linewidth]{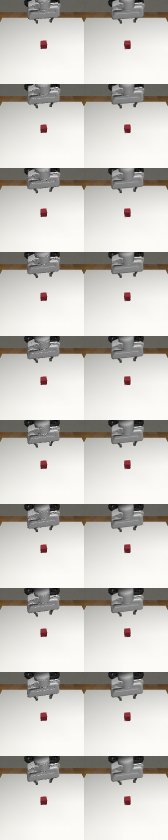}
        
        \includegraphics[width=\linewidth]{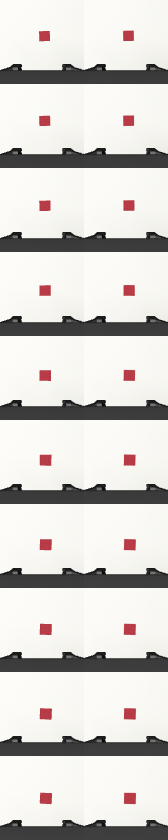}
        \centering{Lift,~mh}
    \end{minipage}
    \begin{minipage}{0.12\textwidth}
        \includegraphics[width=\linewidth]{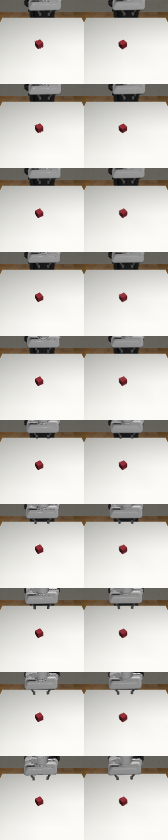}
        \includegraphics[width=\linewidth]{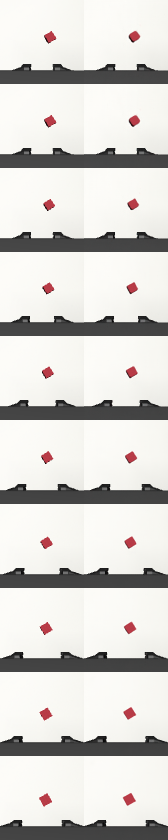}
        \centering{Lift,~ph}
    \end{minipage}
    \begin{minipage}{0.12\textwidth}
        \includegraphics[width=\linewidth]{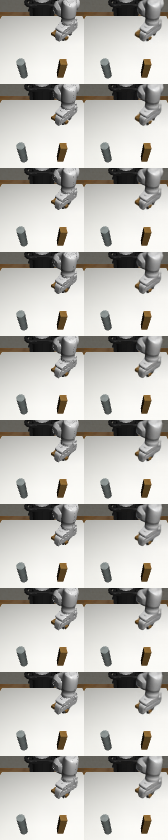}
        \includegraphics[width=\linewidth]{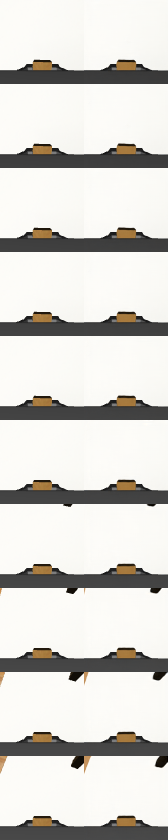}
        \centering{Square,~mh}
    \end{minipage}
    \begin{minipage}{0.12\textwidth}
        \includegraphics[width=\linewidth]{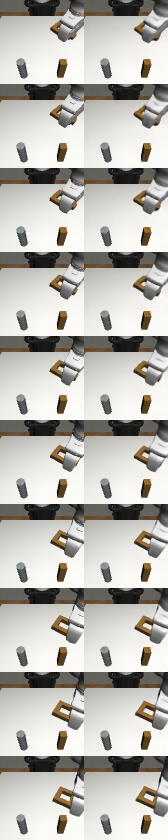}
        \includegraphics[width=\linewidth]{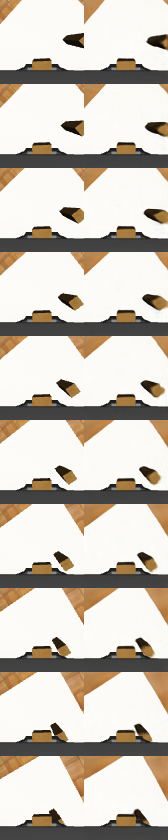}
        \centering{Square,~ph}
    \end{minipage}
    \caption{Reconstruction results on the validation set. For each dataset, left: original image, right: reconstructed image.}
    \label{fig:recon_sample_apdx1}
\end{figure*}

\begin{figure*}
    \centering

    \begin{minipage}{0.12\textwidth}
        \includegraphics[width=\linewidth]{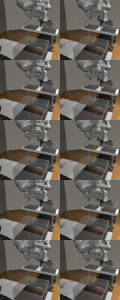}
        \includegraphics[width=\linewidth]{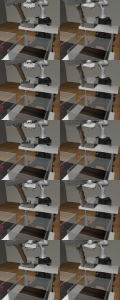}
        \includegraphics[width=\linewidth]{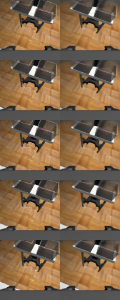}
        \includegraphics[width=\linewidth]{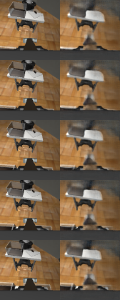}
        \centering{Transport,~mh}
    \end{minipage}
    \begin{minipage}{0.12\textwidth}
        \includegraphics[width=\linewidth]{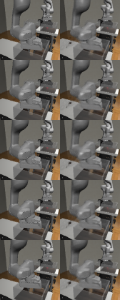}
        \includegraphics[width=\linewidth]{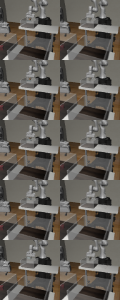}
        \includegraphics[width=\linewidth]{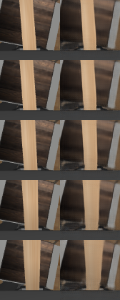}
        \includegraphics[width=\linewidth]{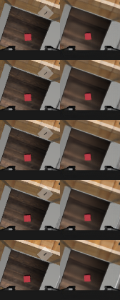}
        \centering{Transport,~ph}
    \end{minipage}
    \begin{minipage}{0.12\textwidth}
        \includegraphics[width=\linewidth]{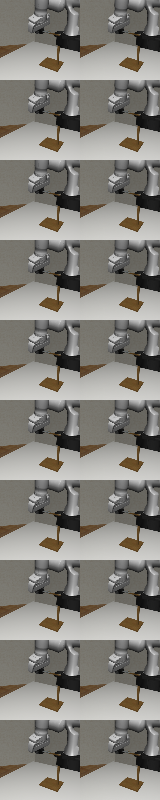}
        \includegraphics[width=\linewidth]{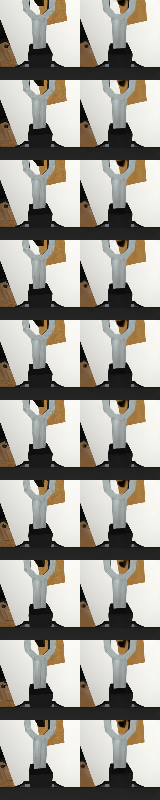}
        \centering{Tool~Hang,~ph}
    \end{minipage}
    \begin{minipage}{0.1197\textwidth}
        \includegraphics[width=\linewidth]{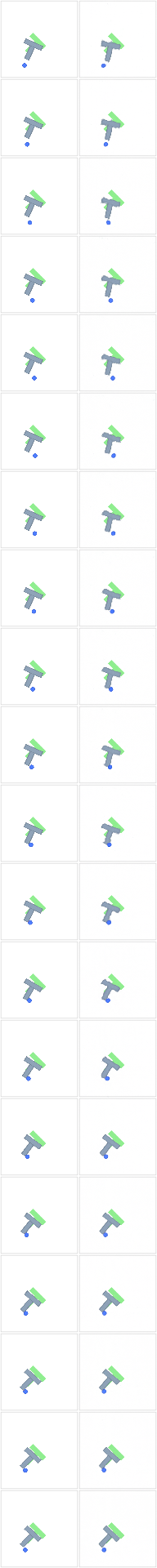}
        \centering{Push-T}
    \end{minipage}
    \begin{minipage}{0.1328\textwidth}
        \includegraphics[width=\linewidth]{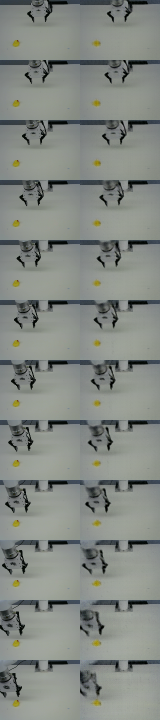}
        \includegraphics[width=\linewidth]{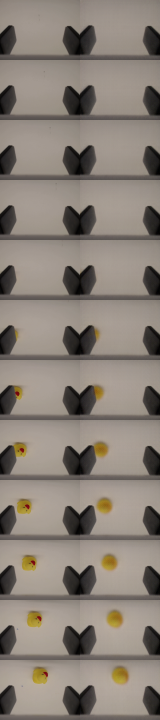}
        \centering{Duck~Lift}
    \end{minipage}
    \begin{minipage}{0.1328\textwidth}
        \includegraphics[width=\linewidth]{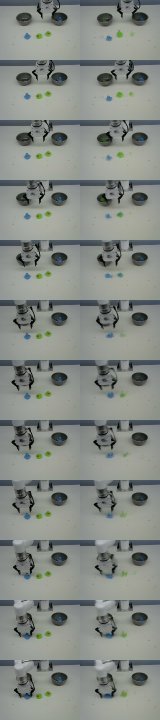}
        \includegraphics[width=\linewidth]{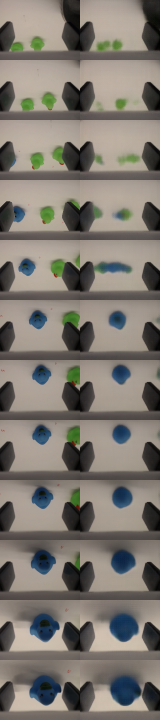}
        \centering{Duck~Collect}
    \end{minipage}
    \caption{Reconstruction results on the validation set. For each dataset, left: original image, right: reconstructed image.}
    \label{fig:recon_sample_apdx2}
\end{figure*}

In Fig.~\ref{fig:recon_sample_apdx1} and Fig.~\ref{fig:recon_sample_apdx2}, we show multiple pairs of original (left) and reconstructed (right) images randomly selected from the validation set. 
For simulated environments, Crossway Diffusion provides surprisingly well-reconstructed images.
For real-world environments, the reconstructions preserve most of the robot and duck structures, while failing on the details (mouth and eyes) of the duck. 
However, we note that bad reconstruction, especially on non-task-related details, does not necessarily mean poor performance. 

\subsection{Example Episodes of Crossway Diffusion}
\label{sec:example_traj}
\begin{figure*}[htp]
    \centering
    \includegraphics[width=0.9\textwidth]{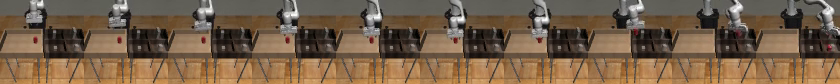}\\
    \centering{Can,~mh (71 steps)}
    \includegraphics[width=0.9\textwidth]{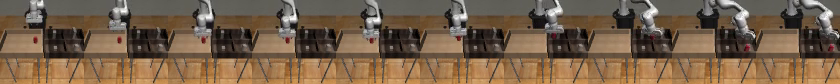}\\
    \centering{Can,~ph (49 steps)}
    \includegraphics[width=0.9\textwidth]{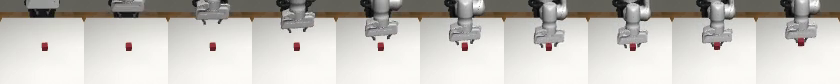}\\
    \centering{Lift,~mh (45 steps)}
    \includegraphics[width=0.9\textwidth]{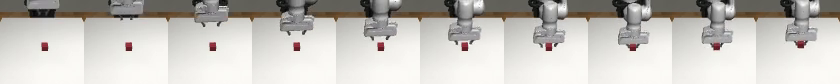}\\
    \centering{Lift,~ph (27 steps)}
    \includegraphics[width=0.9\textwidth]{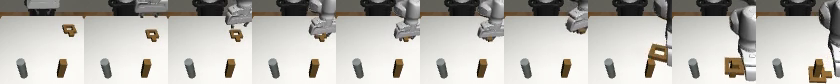}\\
    \centering{Square,~mh (135 steps)}
    \includegraphics[width=0.9\textwidth]{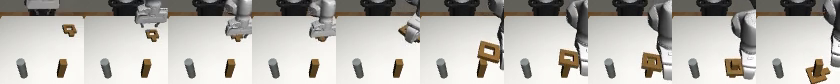}\\
    \centering{Square,~ph (78 steps)}
    \includegraphics[width=0.9\textwidth]{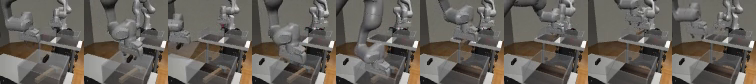}\\
    \centering{Transport,~mh (288 steps)}
    \includegraphics[width=0.9\textwidth]{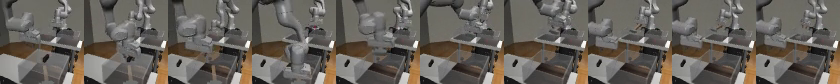}\\
    \centering{Transport,~ph (206 steps)}
    \includegraphics[width=0.9\textwidth]{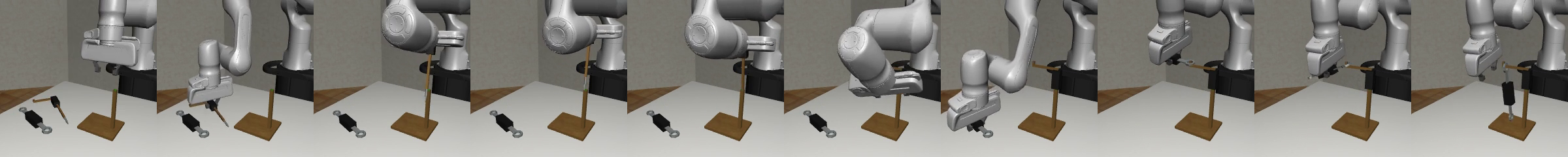}\\
    \centering{Tool~Hang,~ph (198 steps)}
    \includegraphics[width=0.9\textwidth]{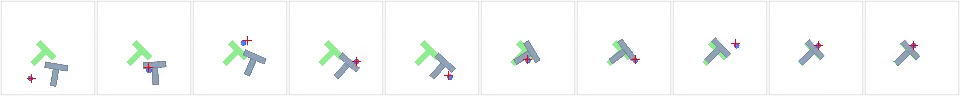}\\
    \centering{Push-T (157 steps)}
    \caption{Example episodes of Crossway Diffusion in simulated environments. The episode length is reported as well.}
    \label{fig:sampletraj}
\end{figure*}

In this section, we present example episodes of Crossway Diffusion on different datasets. In Figure \ref{fig:sampletraj}, for each episode, we sample 10 images with a fixed interval covering the whole episode.

\end{document}